\documentclass[10pt, a4paper]{article}
\usepackage{lrec2022} 
\usepackage{multibib}
\newcites{languageresource}{Language Resources}
\usepackage{graphicx}
\usepackage{tabularx}
\usepackage{soul}
\usepackage{titlesec}
\titleformat{\section}{\normalfont\large\bfseries\center}{\thesection.}{1em}{}
\titleformat{\subsection}{\normalfont\SmallTitleFont\bfseries\raggedright}{\thesubsection.}{1em}{}
\titleformat{\subsubsection}{\normalfont\normalsize\bfseries\raggedright}{\thesubsubsection.}{1em}{}
\renewcommand\thesection{\arabic{section}}
\renewcommand\thesubsection{\thesection.\arabic{subsection}}
\renewcommand\thesubsubsection{\thesubsection.\arabic{subsubsection}}

\usepackage{epstopdf}
\usepackage[utf8]{inputenc}

\usepackage{hyperref}
\usepackage{xstring}

\usepackage{color}

\usepackage[utf8]{inputenc}

\usepackage{microtype}
\usepackage{times}
\usepackage{latexsym}

\usepackage{enumitem}
\usepackage{xspace} 
\usepackage{graphicx}
\usepackage{caption}
\usepackage{subcaption}
\newcommand{\fscore}{\ensuremath{\mathcal{F}}\xspace}
\newcommand{\accuracy}{\ensuremath{\mathcal{A}}\xspace}
\newcommand{\jaccard}{\ensuremath{\mathcal{J}}\xspace}

\newcommand{\arAraBERT}{ar--Arabert}
\newcommand{\enAraBERT}{en--Arabert}
\newcommand{\arenAraBERT}{ar+en--Arabert}

\newcommand{\esBetoBERT}{es--BetoBert}
\newcommand{\enBetoBERT}{en--BetoBert}
\newcommand{\esenBetoBERT}{es+en--BetoBert}

\title{Cross-lingual Emotion Detection}

\name{Sabit Hassan$^{1}$, Shaden Shaar$^{2}$, Kareem Darwish$^{3}$} 

\address{$^{1}$University of Pittsburgh, $^{2}$Cornell University, $^{3}$aiXplain Inc.\\
         $^{1}$Pittsburgh, PA 15260, $^{2}$Ithaca, NY 14850, $^{3}$Los Gatos, CA 95032 \\
         sah259@pitt.edu, ss2753@cornell.edu, kareem.darwish@aixplain.com\\}

\abstract{
Emotion detection can provide us with a window into understanding human behavior. Due to the complex dynamics of human emotions, however, constructing annotated datasets to train automated models can be expensive. Thus, we explore the efficacy of cross-lingual approaches that would use data from a source language to build models for emotion detection in a target language.  
We compare three approaches, namely: i) using inherently multilingual models; ii) translating training data into the target language; and iii) using an automatically tagged parallel corpus. In our study, we consider English as the source language with Arabic and Spanish as target languages. We study the effectiveness of different classification models such as BERT and SVMs trained with different features. Our BERT-based monolingual models that are trained on target language data surpass state-of-the-art (SOTA) by 4\% and 5\% absolute Jaccard score for Arabic and Spanish respectively. Next, we show that using cross-lingual approaches with English data alone, we can achieve more than 90\% and 80\% relative effectiveness of the Arabic and Spanish BERT models respectively. Lastly, we use \textit{LIME} to analyze the challenges of training cross-lingual models for different language pairs.
 \\ \newline  }

\begin{document}

\maketitleabstract
\section{Introduction}
Detecting emotions in text has a wide variety of applications ranging from identifying anger in customer responses to evaluating the emotional well-being of individuals and societies.  Hence, there has been much recent work on automatically detecting emotions in text, with special emphasis on social media posts. Despite contention \cite{ortony1990s}, much work on psychology suggests that \textit{basic} emotions are universal \cite{ekman1999basic,scarantino2011don}.  However, the expressions and perceptions of emotions may exhibit cross-cultural variations \cite{hareli2015cross}, which may complicate cross-lingual learning.  Nonetheless, the ability to perform such cross-lingual classification of emotions effectively is advantageous as it would avoid the laborious task of building emotion detection datasets, which involves collecting representative tweets and properly tagging them. 

While cross-lingual methods have shown success for some tasks such as cross-lingual search \cite{chin2014cross}, named entity recognition \cite{darwish-2013-named}, and sentiment analysis \cite{zhou2016attention}, lesser success has been observed for other tasks such as parsing \cite{guo2015cross} and offensive language detection \cite{pamungkas2019cross}, due to the divergences in linguistic and cultural specificities.  In this paper, we explore the efficacy of using an existing tagged emotion detection corpus from one language to another, and we perform rigorous analysis of results to ascertain the sources of errors. Specifically, we attempt to extend the SemEval2018 English emotion detection dataset \cite{mohammad2018semeval} to Arabic and Spanish. Though the effectiveness of cross-lingual training likely yields lower results compared to training on tagged corpus in the target language, we strive to achieve the lowest possible drop in effectiveness, which would imply that emotions may generalize across linguistic and cultural boundaries. The efficacy of cross-lingual models depends on whether emotions are expressed in similar ways across languages and cultures and on the effectiveness of cross-lingual approaches. In the case of Arabic and English, they are not only different in script, morphology, syntax, etc., they are typically spoken by people from different cultures. \textcolor{black}{Conversely, Spanish shares more in common with English in terms of script and culture. 
}

Given English data, we employ a variety of cross-lingual methods that we compare to using smaller Arabic or Spanish training sets and to mixtures of English with either Arabic or Spanish data. Recent development of inherently multilingual models such as pre-trained multilingual Transformer models, e.g. multilingual BERT (mBERT) \cite{devlin-etal-2019-bert}, and the Multilingual Universal Sentence Encoder (mUSE) \cite{cer2018universal} present new opportunities for cross-lingual approaches. We compare the effectiveness of such models to more traditional models that involve translating the training set into the target language and then retraining.  We also experiment with using a parallel corpus, where we tag the text in the source language, and use the corresponding text in the target language for training. Using the translated training set and the parallel corpus enables us to use more genre specific static embeddings, target language contextual embeddings, and linguistic features to obtain more effective classification, as compared to using mBERT embeddings. In addition, we study combination of the aforementioned approaches. 
To ascertain the efficacy of cross-lingual training we pursued 3 distinct paths as follows:
\begin{enumerate}[leftmargin=*]
\item  using English training data exclusively. To cross the language barrier, we used: multilingual BERT (mBERT), transformer-based multilingual Universal Encoder, automatic machine translation of the training set, and automatic tagging of the English side of a large Arabic-English/Spanish-English parallel corpora and using the Arabic/Spanish sides for training. 
\item using Arabic/Spanish training data exclusively.  
\item using mixtures of Arabic or Spanish in combination with English training data.  In this approach, Arabic or Spanish training sets are augmented with English data.  We crossed the language barrier using the aforementioned methods.
\end{enumerate}
The contributions of this paper are as follows:
\begin{itemize}[leftmargin=*]
    \item We show that cross-lingual models, \textcolor{black}{with English as the source,} can achieve more than 90\% and 80\% of the relative effectiveness of training SOTA monolingual models for Arabic and Spanish respectively. This implies that emotions are generally expressed in similar ways across languages and cultures.  We show that the effectiveness of different cross-lingual approaches follow similar trends for both languages.
    \item \textcolor{black}{As a byproduct of our work, we trained Arabic and Spanish models that notably beat the best SemEval2018 results (52.9 Jaccard vs. 48.9 for Arabic, and 52.4 vs. 46.9 for Spanish)}. 
    \item \textcolor{black}{Using LIME \cite{ribeiro2016lime} for interpretability, we compare models trained on target and source languages. We manually annotate and analyze LIME output to understand the potential challenges of cross-lingual approaches when dealing with different language pairs.}
\end{itemize}

In section \ref{related}, we discuss theories behind human emotions, their usage in the field of natural language processing, and previous work that leverages cross-lingual models. In section \ref{System}, we describe the dataset and systems used in this work and present the results. In section \ref{interpreting}, we identify limitations of cross-lingual models across different cultures and lastly, in section \ref{discussion}, we present a summary of our findings.

\section{Background and Related Work}
\label{related}
Although there is no definitive list of all human emotions, \newcite{ekman1999basic} and \newcite{plutchik1991emotions} suggested that there are 6 and 8 basic emotions respectively. 
\newcite{plutchik1991emotions} further suggested that aside from the 8 primary emotions, there are secondary emotions that arise from combinations of the primary emotions.
\newcite{wang2020review} reviewed different emotion categorization models and listed 65 different emotions that are mentioned in the literature.  The discrepancy between emotion categories is reflected in tagged corpora where some use the basic 6 suggested by \newcite{ekman1999basic} (ex. Abdul-Mageed et al., \shortcite{abdul2016dina}), others use Plutchik's 8 (ex. \cite{alhuzali2018enabling}), and yet others use Pultchik's 8 with secondary emotions, such as love (joy + trust), optimism (anticipation + joy), and pessimism (anticipation + fear) (ex. \cite{mohammad-etal-2018-semeval}).

Although there has been much work on sentiment analysis \cite{mourad-darwish-2013-subjectivity,elmadany2018arsas,hassan-etal-2021-asad}, work on Arabic emotion detection has been relatively limited. There are a few dataset for Arabic emotion detection.  
The datasets of \newcite{abdul2016dina}, \newcite{alkhatib2017emotional}, and \newcite{badarneh2018fine}, which contain 3,000, 10,065, and 11,503 
tweets respectively, were tagged using Ekman's basic 6 emotions.  \newcite{alhuzali2018enabling} used Plutchik's 8 basic emotions to label 7,268 tweets.  The construction of most of the aforementioned datasets was done by searching tweets using words or hashtags that indicate specific emotions and then manually labeling these tweets.  \newcite{hussien2016emoticons} used emoticons to find relevant tweets instead of indicative words. 
Similar to Arabic, there has been much work on Spanish sentiment analysis, but little work Spanish emotion detection \cite{MIRANDA2017REVIEW,segura-bedmar-etal-2017-exploring}.
\newcite{plaza-del-arco-etal-2020-emoevent} created a multilingual corpus consisting of 8.4K Spanish and 7.3K English tweets annotated for Ekman's 6 basic emotions and neutral emotion. Most relevant to our work is the SemEval2018 dataset \cite{mohammad-etal-2018-semeval}, a multilingual publicly available dataset. The data consists of 11K tweets in English, 4.3K in Arabic and 7K tweets in Spanish tagged with one or more of 11 labels corresponding to Plutchik's 8 basic emotions in addition to love, optimism, and pessimism.

Many methods have been used for classifying emotions in text ranging from classical machine learning approaches such as Naive Bayes and Support Vector Machine (SVM) classifiers \cite{alkhatib2017emotional,hussien2016emoticons,PLAZADELARCO-2020-Improved} to deep learning approaches such as recurrent neural networks \cite{abdullah2018teamuncc,alhuzali2018enabling}.  Work on cross-lingual emotion detection is relatively scant.  \newcite{ren2018emotion} used a \mbox{biLSTM} model to classify emotions in multilingual text.  \newcite{ahmad2020borrow} used a multilingual embedding space to improve Hindi emotion detection by leveraging an English dataset. Aside from emotion detection, cross-lingual learning was employed by many for sentiment analysis.  For example, 
\newcite{dong-de-melo-2019-robust} used self-learning with multilingual BERT for document classification and Chinese sentiment analysis. Similarly,  \newcite{Xu-Cross-2017} used a parallel corpus with adversarial feature adaptation to perform cross-lingual text classification with English as a source language and German, French, Japan and Chinese as target languages. \newcite{jebbara-cimiano-2019-zero} use multilingual word embeddings with convolutional neural network for extracting opinion target expressions. \newcite{Abdullah-SEDAT} translated Arabic to English to classify Arabic sentiment using an English model.  Further, \newcite{bel-cross-2003} and \newcite{mihalcea-etal-2007-learning} used bilingual dictionaries to translate sentiment words and documents.  
More recent efforts used automated Machine Translation (MT) instead of bilingual dictionaries \cite{wan-2009-co}. \newcite{klinger-cimiano-2015-instance} used statistical (MT) for sentiment analysis in German/English pair. Translation has also been applied to other tasks such as semantic role labeling \cite{fei-etal-2020-cross} and dependency parsing \cite{zhang-etal-2019-cross}. 
\section{System Setup}
\label{System}
\subsection{Datasets} 
We used the SemEval2018 emotion detection dataset (Affect in Tweets: Task E-c) to evaluate our cross-lingual approaches \cite{mohammad-etal-2018-semeval}. Table \ref{tab:distribution} lists distribution of tweets across the different languages and train-dev-test splits.
\begin{table}[!h]
\centering
\small
\begin{tabular}{lccc}
\hline \textbf{Language} & \textbf{Train} & \textbf{Dev} &\textbf{Test}\\ \hline
English &  6,838 & 886 & 3,259\\
Arabic & 2,278 & 585 & 1,518\\
Spanish & 3,561 & 679 & 2,854\\
\hline
\end{tabular}
\caption{\label{tab:distribution} Distribution of tweets in SemEval2018 dataset}
\end{table}

Each tweet was manually annotated, using crowd-sourcing, for the presence of 11 different emotions: joy, sadness, anger, fear, disgust, surprise, trust, anticipation, love, optimism, and pessimism.  
Further, given the English training and development sets, we translated them into Arabic and Spanish using Google translation APIs to be used with our translation approach. 

For our approach requiring a parallel corpus for Arabic, we used the dataset of \newcite{mubarak-etal-2020-constructing}, which consists of 166K pairs of Arabic-English parallel tweet\textemdash pairs that contain the same content but written separately in Arabic and English. 
The advantage of using parallel tweets instead of translating the SemEval2018 dataset (using MT) is that parallel tweets resulted from human translation, which would hopefully better transfer meaning and emotions. 
However, the parallel tweets are not annotated for emotions. To create the annotations, 
we tagged the English side using our best setup, namely BERT-uncased, and given sentences 
that were tagged with at least 3 emotions on the English side, we transferred the tags to the corresponding parallel Arabic sentences.  Placing such conditions would improve our chances of getting more accurately 
tagged sentences. We experimented with different filters, and retaining tweets with more than 3 emotions led to the best results.  We omitted these experiments for conciseness. After applying our filter, we end up with 4,450 pairs of tagged parallel tweets.

For Spanish, due to lack of a parallel tweets corpus, we opted to use the Open Subtitles dataset \cite{TIEDEMANN2012Parallel}, which has millions of parallel movie subtitles in different languages. However, to be comparable to Arabic, we randomly pick 166K pairs to have the same starting number. Then we apply the similar filters as Arabic and ended up with 20K pairs, suggesting that movie subtitles have more emotions than parallel tweets collected using the method of \newcite{mubarak-etal-2020-constructing}. 


\subsection{Models}
\textbf{Transformer Model:} Transformer-based pre-trained contextual embeddings, such as BERT \cite{devlin-etal-2019-bert}, ULMFIT~\cite{howard-ruder-2018}, and OpenAI GPT ~\cite{radford2018} have led to large improvements in many NLP tasks.  \textcolor{black}{We used such transformer based models, namely bert-base-uncased (BERT-uncased), trained on lower cased English text, bert-base-multilingual-uncased (mBERT), which is trained on Wikipedia text from different languages, including Arabic, Spanish, and English, AraBERT \cite{Antoun2020AraBERT}, which is trained on a larger Arabic news corpus that contains roughly 2.5B tokens, and, BetoBERT \cite{CaneteCFP2020Spanish}, which is trained with variety of Spanish corpora consisting of 3B tokens.}
All the aforementioned models are pre-trained on identical architectures, namely an encoder with $12$ Transformer blocks, a hidden size of $768$, and $12$ self-attention heads. All use SentencePiece segmentation (BP).  We used the HuggingFace implementation\footnote{https://huggingface.co}, and we fine-tuned the contextual embeddings using either English, Arabic, Spanish or a combination of Arabic or Spanish with English training data for 10 epochs with learning rate of 2e\textsuperscript{-5} and batch size of 8. \textcolor{black}{We placed an output layer with 11 output nodes corresponding to the probabilities of each emotion class.}\\
\textbf{Support Vector Machines (SVMs):}  SVMs have been shown to be effective for a variety of text classification tasks \cite{forman2008bns,mubarak-hassan-2021-ul2c,abdelali-etal-2021-qadi}. \textcolor{black}{Due to the fact neither mBERT or AraBERT were trained on dialectal or social media texts, using SVMs would hopefully allow us to perform feature engineering to achieve competitive results.}  We used the libSVM implementation in scikit learn\footnote{https://scikit-learn.org} with a linear kernel.  We trained a binary classifier for each of the emotions \textcolor{black}{ in a one vs. rest setting}.\\
\textbf{Multi-Layer Perceptron (MLP):}  We implemented an MLP classifier using Keras \cite{chollet2015keras} with a tensorflow backend \cite{tensorflow2015-whitepaper}.  We used it only with mUSE embeddings.  The network had 512 input nodes, two hidden layers with 256 and 128 nodes respectively, and 11 output nodes corresponding to the different emotions.  One of the major advantages of this setup is that we can jointly train using all the labels.  We trained the classifier for a maximum of 50 epochs with early stopping with patience of 3 iterations.
\subsection{Features}
\textbf{Word unigrams:} \textcolor{black}{Training a bag-of-words model often serves as a baseline model (e.g., \cite{mohammad2018semeval}). We weighed word unigrams using term frequency-inverse term document frequency (\textit{tf-idf} weighting) and used with SVMs only.}\\
\textbf{Character $n$-gram:}  Character $n$-grams have been shown to be effective in representing text in retrieval and classification tasks \cite{mcnamee2004character,mubarak2020spam}. \textcolor{black}{Using them helps overcome some of the effects of: the frequent creative word spellings on social media; and the complex derivational morphology of Arabic -- availing the need for stemming.} \textcolor{black}{Similar to word unigrams, we used character $n$-grams as features when using an SVM classifier only, and we used \textit{tf-idf} weighting. }\\
\textbf{Word Embeddings:} For Arabic, we used Mazajak static word-level skip-gram embeddings \cite{abu-farha-magdy-2019-mazajak}, which were trained on 250M Arabic tweets with 300-dimensional vectors.  Since they were trained on tweets, they have led to improved tweet text classification results, compared to using mBERT, for some tasks \cite{mubarak2020arabic,hassan-etal-2020-alt}. 
Due to unavailability of Twitter specific embeddings for Spanish\footnote{Spanish embeddings based on tweets \url{https://www.spinningbytes.com/resources/wordembeddings/} are unavailable}, we opted for Spanish Billion Word (SBW) embeddings \cite{cardellinoSBWCE}, trained with approximately 1.5 Billion words from different corpora including books, newspapers and Wikipedia dumps.\\
\textbf{mUSE:} mUSE is a transformer based model that is pre-trained on multiple tasks simultaneously \cite{cer2018universal}.  The models use BP segmented text and produces a 512 feature vector for each input sentence, and it does not require fine-tuning.  

\subsection{Experimental Setup}
\paragraph{Evaluation Metrics:} 
we compared results using the average Jaccard similarity (\jaccard) between the predicted labels and the ground truth. Jaccard score is defined as the size of the intersection divided by the size of the union of two label sets. It is effective in comparing results in a multi-label scenario and is the official metric for the SemEval2018 emotion detection task. We also report the macro F-measure across all labels (\fscore) and the average accuracy (\accuracy). All results are reported on the official test sets of SemEval2018 emotion detection task. 
\paragraph{Baseline Monolingual Models: }
Initially, we opted to produce SOTA models trained on Arabic or Spanish data exclusively to compare cross-lingual models to. We trained several Arabic and Spanish models using the SemEval2018 emotion detection dataset: 1)
    SVM models trained using: word uni-grams (W[1-1]), character $n$-grams (C[1-6]), with $n$ ranging from 1 to 6; Mazajak/SBW embeddings; character $n$-grams + Mazajak/SBW; and mUSE embeddings; 2)
    an MLP model trained on mUSE embeddings; 3)
    fine-tuned mBERT, AraBERT, or BetoBERT models.

Table \ref{tab:ArabicBaseline} reports on the Arabic baseline results. As the results show, using AraBERT yielded the best results with a \jaccard score of 52.9. It is noteworthy that the best result reported in the SemEval2018 shared task was 48.9 \cite{mohammad-etal-2018-semeval}.  Using character $n$-grams with Mazajak embeddings performed better than using either mUSE or mBERT.  This is consistent with results obtained for other classifications tasks involving Arabic tweets \cite{mubarak2020arabic,hassan-etal-2020-alt-semeval}. 

Table \ref{tab:SpanishBaseline} reports on the Spanish baseline results.  Using BetoBERT yielded the best results with a \jaccard score of 52.4, while the best result reported in the SemEval2018 shared task was 46.9 \cite{mohammad-etal-2018-semeval}.  In contrast to Arabic, use of word embeddings (SBW) did not result in better results compared to mUSE or mBERT. This could be due to the fact that SBW are not Twitter specific embeddings, as opposed to Mazajak. 
\textcolor{black}{Due to poor performances of using word-unigrams, we excluded them from further experiments.}


\begin{table}
\centering
\small
\begin{tabular}{lcccc}
\hline \textbf{Model} & \textbf{Features} & \textbf{\jaccard} & \textbf{\fscore} & \textbf{\accuracy}\\ \hline
\textit{SemEval} & & 48.9 & 46.1 & -\\
SVM & W[1-1] & 35.5 & 34.3 & 83.3\\
SVM & C[1-6] & 42.0 & 37.9 & 85.3\\
SVM & Mazajak & 46.3 & 44.3 & 84.7\\
SVM & \small{C[1-6]+Mazajak} & 48.6 & 46.2 & 85.5\\
SVM & mUSE & 38.9 & 38.5 & 82.9\\
MLP & mUSE & 40.3 & 37.0 & 83.3\\
mBERT & - & 45.3 &	41.4 &	84.2\\
AraBERT & - & \textbf{52.9}	& \textbf{48.9} & \textbf{86.6}\\
\hline
\end{tabular}
\caption{\label{tab:ArabicBaseline} Models trained on Arabic data. SemEval refers to the highest ranked system in the shared task.}
\end{table}

\begin{table}
\centering
\small
\begin{tabular}{lcccc}
\hline \textbf{Model} & \textbf{Features} & \textbf{\jaccard} & \textbf{\fscore} & \textbf{\accuracy}\\ \hline
\textit{SemEval} & & 46.9 & 40.7 & -\\
SVM & W[1-1] & 29.7 & 28.0 & 87.1\\
SVM & C[1-6] & 35.0 & 31.3 & 88.4\\
SVM & SBW & 22.9 & 20.2 & 85.8\\
SVM & \small{C[1-6]+SBW} & 37.4	& 35.2 & 87.9\\
SVM & mUSE & 39.4 &	40.0 & 88.2\\
MLP & mUSE & 42.3 & 41.5 & 88.6\\
mBERT & - & 43.7 & 35.8 & 88.5\\
BetoBERT & - & \textbf{52.4} & \textbf{53.7} & \textbf{89.8}\\
\hline
\end{tabular}
\caption{\label{tab:SpanishBaseline} Models trained on Spanish data. SemEval refers to the highest ranked system in the shared task.}
\end{table}
\begin{table}[h]
\centering
\small
\begin{tabular}{lcccc}
\hline \textbf{Model} & \textbf{Features} & \textbf{\jaccard} &\textbf{\fscore} & \textbf{\accuracy}\\ \hline
\textit{SemEval} & - & \textbf{58.8} & \textbf{52.1} & - \\
SVM & W[1-1] & 38.8 & 37.2  & 83.0\\
SVM & C[1-6] & 43.1 & 39.5  & 84.6\\
SVM & mUSE & 48.4 & 43.7 & 85.6\\
MLP & mUSE & 51.3 &	45.4 & 85.7\\
mBERT & - & 54.6 & 46.1 & 87.0\\
BERT-uncased & - & 56.4 & 53.9 & 87.3\\
\hline
\end{tabular}
\caption{\label{tab:bestEnglishResults} Models for English emotion analysis. SemEval refers to the highest ranked system in the shared task.}
\end{table}
\paragraph{Cross-Lingual Models: } 
\begin{table*}[t]
\centering
\small
\setlength\tabcolsep{3.2pt}
\begin{tabular}{@{}lcc|ccc|ccc@{}}
\multicolumn{3}{c}{} & \multicolumn{3}{|c}{Cross-lingual} & \multicolumn{3}{|c}{Combined} \\ \hline
\textbf{Approach} & \textbf{Model} & \textbf{Features} & \textbf{\jaccard} & \textbf{\fscore} & \textbf{\accuracy} & \textbf{\jaccard} & \textbf{\fscore} & \textbf{\accuracy} \\
\hline
\multicolumn{3}{l}{Best Arabic baseline result} & \textbf{52.9}	& \textbf{48.9} & \textbf{86.6} & \multicolumn{3}{c}{} \\ 
\hline
M 	&	 SVM 	&	 mUSE 	&	32.4	&	31.6	&	79.7	& 37.9 & 37.0 & 83.1\\
M 	&	 MLP 	&	 mUSE 	&	35.8    &	33.4	&   79.9	& 41.8 & 40.5 & 83.0\\
M 	&	 mBERT 	&	 - 	&	20.5    &	19.5    &   	76.4	& 46.7 & 44.6 & 83.8\\
\hline											
T 	&	 SVM 	&	 C[1-6] 	&	28.0	&	27.3	&	81.0	& 44.4 & 41.3 & 85.4\\
T 	&	 SVM 	&	 Mzjk 	&	39.3	&	37.1	&	82.8	& 43.7 & 40.4 & 85.2\\
T 	&	 SVM 	&	 C[1-6]+Mzjk 	&	42.5	&	40.2	&	82.5	& 48.0 & 45.0 & 85.6\\
T 	&	 AraBERT 	&	 - 	&	\textbf{48.1}   &	\textbf{46.3}   &	\textbf{83.8}	& \textbf{54.1} & \textbf{50.8} & \textbf{86.2}\\
\hline											
P 	&	 SVM 	&	 C[1-6] 	&	30.4	&	24.9	&	74.4	& 44.1 & 40.5 & 85.3\\
P 	&	 SVM 	&	 Mzjk 	&	36.5	&	31.9	&	74.6	& 45.1 & 43.3 & 84.3\\
P 	&	 SVM 	&	 C[1-6]+Mzjk 	&	37.5	&	32.4	&	75.2	& 48.4 & 46.2 & 85.3\\
P 	&	 AraBERT 	&	 - 	&	36.7	&	31.6	&	75.2	& 53.1 & 50.5 & 85.9\\
\hline											
M + T	&	 SVM 	&	 mUSE 	&	34.3	&	32.5	&	80.5 & 38.4 & 36.3 & 83.3\\ 
M + T	&	 MLP 	&	 mUSE 	&	36.8    &	34.6    &	80.7 & 41.5 & 38.6 & 83.0\\
M + T	&	 mBERT 	&	 -  &   36.0   &	36.0    &	79.9 & 45.9 & 44.0 & 83.6\\
\hline
P + M &   SVM     &   mUSE    &   36.7    &	33.9    &	78.9	& 39.1 & 37.4 & 82.8\\
P + M &   MLP     &   mUSE    &   37.9    &	34.9    &	79.2	& 41.4 & 38.6 & 82.8\\
P + M &   mBERT   &   -   &  30.2     &	26.1    &	70.6    	& 46.0 & 44.2 & 83.3\\
\hline
P + T  &   SVM     &   C[1-6]  &   35.6    &	34.2    &	81.2 & 45.4 & 41.9 & 85.1\\
P + T  &   SVM     &   Mzjk  &  40.5    &	37.6    &	81.8 & 43.0 & 40.4 & 84.9\\
P + T  &   SVM     &   C[1-6]+Mzjk  &     44.0    &	40.1    &	82.2 & 48.0 & 45.1 & 85.5\\
P + T  &   AraBERT     &   -  &    47.7    &	45.0    &	82.4 & 53.1 & 50.4 & 85.8\\
\hline
M + P + T   &   SVM     &   mUSE    & 37.2    &	34.8    &	80.3 & 38.1 & 36.2 & 82.7\\
M + P + T   &   MLP     &   mUSE    & 39.3    &	35.1    &	79.8 & 41.5 & 38.5 & 82.2\\
M + P + T   &   mBERT   &   -   &   38.5    &	36.7    &	78.8 & 46.4 & 44.5 & 83.6\\
\hline
\hline
\end{tabular}
\caption{Arabic cross-lingual and multilingual results. M: Multilingual Model; T: Translation; and P: Parallel corpus; Mzjk: Mazajak embeddings}
\label{tab:cross-lingual-results}
\end{table*}

\begin{table*}[!t]
\centering
\small
\setlength\tabcolsep{3.2pt}
\begin{tabular}{@{}lcc|ccc|ccc@{}}
\multicolumn{3}{c}{} & \multicolumn{3}{|c}{Cross-lingual} & \multicolumn{3}{|c}{Combined} \\ \hline
\textbf{Approach} & \textbf{Model} & \textbf{Features} & \textbf{\jaccard} & \textbf{\fscore} & \textbf{\accuracy} & \textbf{\jaccard} & \textbf{\fscore} & \textbf{\accuracy} \\
\hline
\multicolumn{3}{l}{Best Spanish baseline result} & \textbf{52.4} & \textbf{53.7} & \textbf{89.8} & \multicolumn{3}{c}{} \\ 
\hline
M 	&	 SVM 	&	 mUSE 	& 36.6 & 36.9 & 85.7 & 38.8 & 37.4 & 88.4\\
M 	&	 MLP 	&	 mUSE 	& 37.4 & 38.7 &	84.2 & 42.5 & 41.4 & 88.4\\
M 	&	 mBERT 	&	 - 	&	26.5 & 23.6 & 83.3 & 42.3 & 32.8 & 88.8\\
\hline											
T 	&	 SVM 	&	 C[1-6] 	& 30.1 & 28.3 & 85.3 & 38.4 & 37.0 & 88.3\\
T 	&	 SVM 	&	 SBW 	&	23.4 & 21.9 & 83.0 & 22.0 & 19.6 & 85.6\\
T 	&	 SVM 	&	 C[1-6]+SBW 	& 31.5 & 30.3 & 84.1 & 38.9 & 37.7 & 87.9\\
T 	&	 BetoBERT 	&	 - 	&	\textbf{42.3} & \textbf{44.1} & \textbf{85.5} & \textit{51.8} & \textit{53.4} & \textit{89.2}\\
\hline											
P 	&	 SVM 	&	 C[1-6] 	& 27.1 & 31.8 & 76.6 & 37.9	& 38.5 & 86.2\\
P 	&	 SVM 	&	 SBW 	&	24.3 & 24.8 & 78.5 & 23.3 & 23.6 & 83.2\\
P 	&	 SVM 	&	 C[1-6]+SBW 	& 27.9 & 32.8 & 76.5 & 38.6 & 39.3 & 86.1\\
P 	&	 BetoBERT 	&	 - 	&	34.1 & 40.5	& 77.4 & 50.5 & 51.7 & 88.2\\
\hline											
M + T	&	 SVM 	&	 mUSE 	& 36.8 & 35.9 & 85.3 & 39.6	& 38.2 & 88.2\\ 
M + T	&	 MLP 	&	 mUSE 	& 37.6 & 36.4 & 84.8 & 42.0 & 42.0 & 87.8\\
M + T	&	 mBERT 	&	 -  &   37.0 & 33.9 & 84.6 & 42.2 & 34.0 & 88.4\\
\hline
P + M &   SVM     &   mUSE    & 36.1 & 38.1 & 82.4 & 36.9 & 37.4 & 86.7\\
P + M &   MLP     &   mUSE    & 37.0 & 39.4 & 82.1 & 40.1 & 40.8 & 87.2\\
P + M &   mBERT   &   -   &  32.6 & 38.6 & 77.0 & 47.7 & 47.6 & 87.3\\
\hline
P + T  &   SVM     &   C[1-6]  & 33.1 & 35.6 & 81.6 & 40.0 & 40.0 & 86.4\\
P + T  &   SVM     &   SBW  &  24.0 & 24.3 & 81.3 & 23.4 & 23.3 & 83.6\\
P + T  &   SVM     &   C[1-6]+SBW  & 33.6 & 36.2 & 81.6 & 39.7 & 40.4 & 86.3\\
P + T  &   BetoBERT     &   -  & 40.7 & 44.9 & 82.3 & 50.2 & 52.5 & 88.0\\
\hline
M + P + T   &   SVM     &   mUSE    & 35.4 & 37.2 & 83.8 & 37.2 & 37.9 & 86.7\\
M + P + T   &   MLP     &   mUSE    & 37.3 & 39.0 & 82.9 & 42.1 & 41.9 & 86.7\\
M + P + T   &   mBERT   &   -   &   38.7 & 41.3 & 82.6 & 47.3 & 47.3 & 87.2\\
\hline
\hline
\end{tabular}
\caption{Spanish cross-lingual and multilingual results. M: Multilingual Model; T: Translation; and P: Parallel corpus; SBW: Spanish Billion Words}
\label{tab:cross-lingual-results-spanish}
\end{table*}
We constructed a large number of experimental setups involving the use of English training data to tag the Arabic and Spanish test set.  For setups involving translation, we used either the training set after translation using MT or the Arabic and Spanish sides of the parallel corpora. 
To use parallel data as a conduit for translation, we experimented with several setups for English emotion detection on the SemEval data. They included models that were similar to the Arabic or Spanish baseline models listed earlier, save ones with Arabic or Spanish specific static embeddings (Mazajak, SBW) \textcolor{black}{and instead of AraBERT or BetoBERT, we used BERT-uncased. The results are summarized in Table \ref{tab:bestEnglishResults}. BERT-uncased outperformed  MLP, SVM and mBERT. Thus, we used it to tag the English side of the parallel corpus. }

As with the baseline setups, after translation we: 1) 
trained an SVM classifiers using: character $n$-grams, where $n$ ranged between 1 and 6 (C[1-6]); Mazajak/SBW embdeddings; or a combination of the character $n$-gram and Mazajak/SBW; and 2)
fine-tuned AraBERT/BetoBERT contextual embeddings, as they performed notably better than mBERT in monolingual Arabic/Spanish baselines. 

We also had two other setups that did not involve translation, namely: using mUSE, where we produced embedding representations of the English training set and the Arabic/Spanish test set, and we used MLP or SVM classifiers; and using the mBERT model, fine-tuned on the English training set, to tag the Arabic/Spanish test set directly.  
We conducted experiments involving other combinations of using multilingual representations (mUSE and mBERT), the translated training set, and the Arabic/Spanish side of the automatically tagged parallel corpus using similar set of models. 

Table \ref{tab:cross-lingual-results} summarizes the results of all cross-lingual approaches using only English data when Arabic is the target language. We observe that mBERT was unable to produce good results. On the other hand, using mUSE to train SVM and MLP classifiers notably outperformed mBERT. 
Using a parallel corpus allowed us to use language specific features, such as character $n$-gram, and we noticed an improvement over using multilingual models. 
However, we suspect that the problem with using parallel text is that tweets are tagged using an imperfect model, and the errors percolate from English to the target language.
Automatically translating the English training set to Arabic led to the best results, with AraBERT achieving a notably higher \jaccard score of 48.1.   
A less notable improvement was noticed when using mUSE, Mazajak embedding, and character $n$-grams on translations compared to parallel text. 
Our attempts to combine translations with parallel text and multilingual representations did not yield results that were better than using translations alone. 
The best overall cross-lingual results that we obtained using AraBERT (\jaccard=48.1) achieved 90.9\% of the best results that were obtained when training on the Arabic data using AraBERT (\jaccard=52.9) and were 98.4\% of the best team's results at the SemEval2018 Arabic emotion detection shared task (\jaccard=48.9). 
We can conclude that the use of translation along with fine-tuned AraBERT model can lead to competitive results availing the need for target language annotation. 

Table \ref{tab:cross-lingual-results-spanish} summarizes the results of all cross-lingual approaches using only English data when target language is Spanish. We observe several similar trends to Arabic results. Among inherently multilingual models, mBERT was outperformed by SVM and MLP trained on mUSE. Translation with transformer model, BetoBERT (\jaccard=42.3), once again produced the best results, and we also see that combination of approaches did not affect results notably.  Similarly, the use of parallel corpora yielded suboptimal results for both languages. 
One difference is the lower relative performance of translation with transformer models for Spanish compared to Arabic. Although it still produces the best results compared to other approaches, its relative efficacy is lower than we observed in the case of Arabic (80.7\% as opposed to 90.9\%). We examine this in more details in the following section.

\paragraph{Combined Models: }
We wanted to see if cross-lingual English training data in conjunction with Arabic/Spanish training data would lead to improved results.  To do so, we followed the same cross-lingual approaches as in the previous subsection. 
The only difference is that we trained the models using both English training data and the Arabic or Spanish data.
For combined Arabic+English models (shown in Table \ref{tab:cross-lingual-results}), we observed similar trends to those observed for cross-lingual results.  Fine-tuning AraBERT on the translated English set in combination with the Arabic set led to the best results.  Combining translations with parallel data and/or multilingual models led to lower results.  While the best results are notably better than training using just the English data, with 6.0 points absolute improvement, the difference compared to using the Arabic training set alone is less pronounced, with a 1.2 points absolute improvement. 
As for Spanish+English models in Table \ref{tab:cross-lingual-results-spanish}, we see similar trends to those for Arabic. These trends show the translation approach outperforming others and there is little difference in effectiveness compared to training on Spanish alone. 
\paragraph{Class Wise Evaluation:}
Lastly, Table \ref{tab:Comparison} compares the per emotion F1-measure when using monolingual, cross-lingual, and combined models.  As can be seen, in the case of Arabic, the monolingual data produced very poor results for some emotions such as \textit{surprise}.  
In the case of Spanish, the cross-lingual data performed poorly for some classes such as \textit{trust}. In both cases, combined models can attenuate this effect to an extent.
\begin{table*}[!t]
\centering
\small
\begin{tabular}{lccccccccccc}
\textbf{Model} & \rotatebox{90}{\textbf{anger}}	& \rotatebox{90}{\textbf{anticipation}}	& \rotatebox{90}{\textbf{disgust}}	& \rotatebox{90}{\textbf{fear}}	& \rotatebox{90}{\textbf{joy}} &	\rotatebox{90}{\textbf{love}} &	\rotatebox{90}{\textbf{optimism}} &	\rotatebox{90}{\textbf{pessimism}}	& \rotatebox{90}{\textbf{sadness}}	& \rotatebox{90}{\textbf{surprise}}	& \rotatebox{90}{\textbf{trust}}\\
\hline
\arAraBERT  & 77.2 & 3.6 &	49.0 & 71.9 & 79.9 & 71.8 &	71.8 & 37.0 & 70.9 & 0.0 & 4.7\\
\enAraBERT & 67.9	&	5.9	&	50.5	&	66.1	&	74.2	&	56.8	&	67.7	&	23.9	&	69.8	&	16.0	&	10.9 \\
\arenAraBERT & 74.9 &	16.4 &	47.5 &	70.3 &	80.4 &	70.3 &	72.7 &	31.8 &	73.7 &	9.3 &	8.9 \\
\hline
\esBetoBERT  & 74.8 & 49.5 & 46.8 & 68.6 & 80.6 & 65.4 & 39.1 & 45.1 & 65.8 & 26.3 & 28.2 \\
\enBetoBERT & 71.9 & 25.5 & 44.1 & 63.4 & 75.0 & 55.1 & 42.4 & 26.7 & 60.4 & 13.7 & 7.3 \\
\esenBetoBERT & 73.4 & 43.4 & 47.0 & 69.8 & 80.8 & 65.9 & 44.0 & 42.6 & 66.4 & 23.7 & 30.6 \\
\hline
\end{tabular}
\caption{Comparison of class-level f1}
\label{tab:Comparison}
\end{table*}
\begin{figure}
    \centering
    \includegraphics[width=0.48\textwidth]{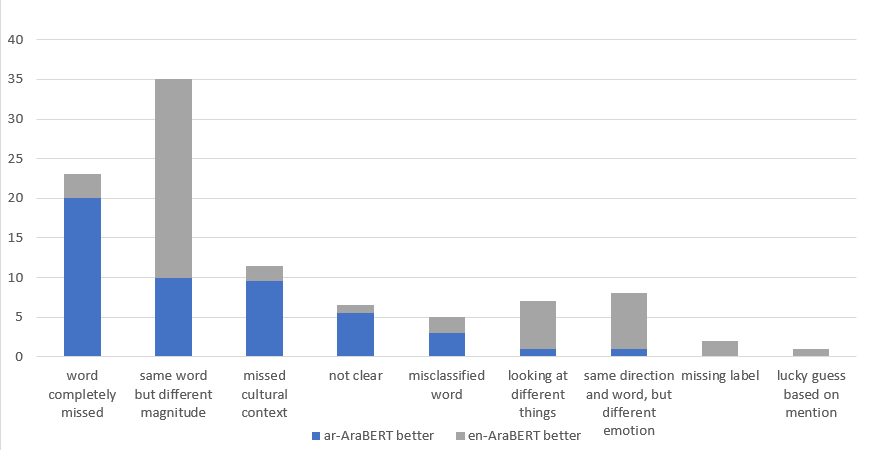}
    \caption{Top error types for Arabic}
    \label{fig:arabicErrorTypes}
\end{figure}
\begin{figure}
    \centering
    \includegraphics[width=0.48\textwidth]{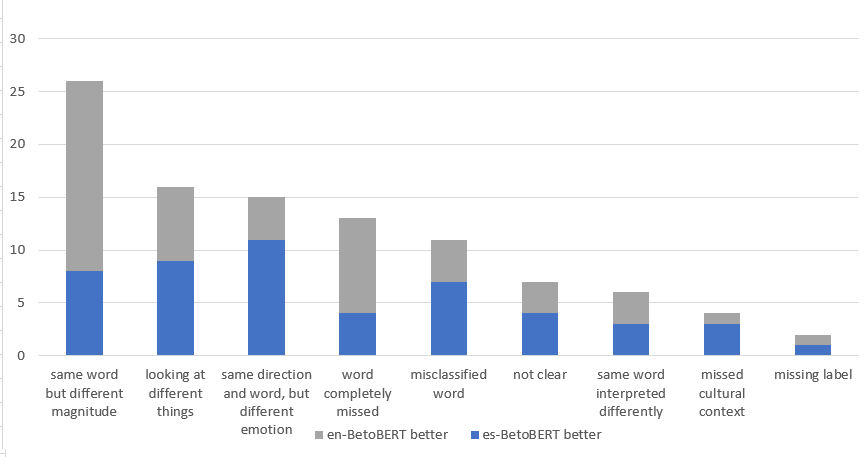}
    \caption{Top error types for Spanish}
    \label{fig:spanishErrorTypes}
\end{figure}
\section{Interpreting Cross-Lingual Differences}
\label{interpreting}
As shown in the experiments in the previous section, the best cross-lingual model for emotions detection were contextual embedding models that are fine-tuned on the translated training set. In this section, we analyze how they compare to monolingual models. To aid our analysis, we used LIME \cite{ribeiro2016lime} to identify the words and phrases that triggered the models to choose certain labels. LIME is an algorithm that tries to determine how specific words affect the results of the classifier by creating variations of an input sentence through randomly removing words. Since this is a multi-label task, each label was treated independently, and a word would contribute differently to each of the labels.  To ascertain the effect of each word on a label, LIME would look at all the variant sentences that contain that word, then average the product of the weight (percentage of words used from the input sentence) and the probability (classifier's score for the variant sentence for a given label).
The higher the score for a word, the higher its impact on the classifier's prediction.

\begin{figure}[pt]
    \centering
    \begin{subfigure}{.28\textwidth}
      \centering
      \includegraphics[scale=0.35]{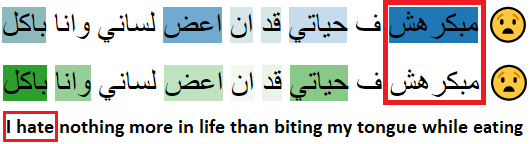}  
      \caption{Word completely missed}
      \label{fig:BetterAr1}
    \end{subfigure}
    \begin{subfigure}{.28\textwidth}
      \centering
      \includegraphics[scale=0.35]{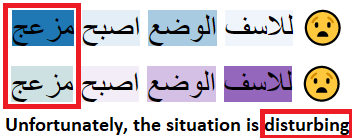}  
      \caption{Same words weighted differently}
      \label{fig:BetterAr2}
    \end{subfigure}
    \begin{subfigure}{.28\textwidth}
      \centering
      \includegraphics[scale=0.35]{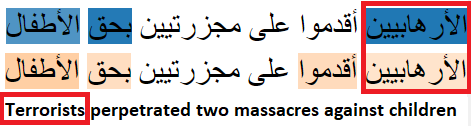}
      \caption{Missed cultural context}
      \label{fig:BetterAr3}
    \end{subfigure}    
    \caption{Examples of monolingual \arAraBERT~outperforming cross-lingual \enAraBERT}
    \label{fig:BetterAr}

\end{figure}

\begin{figure}[ht]
    \centering
    \begin{subfigure}{.4\textwidth}
      \centering
      \includegraphics[scale=0.35]{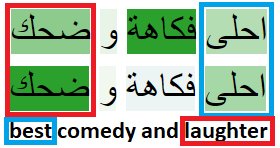}  
      \caption{Same words weighted differently}
      \label{fig:BetterEn1}
    \end{subfigure}
    \begin{subfigure}{.4\textwidth}
      \centering
      \includegraphics[scale=0.35]{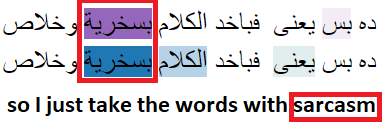}  
      \caption{Same polarization but different emotions}
      \label{fig:BetterEn2}
    \end{subfigure}
    \begin{subfigure}{.4\textwidth}
      \centering
      \includegraphics[scale=0.25]{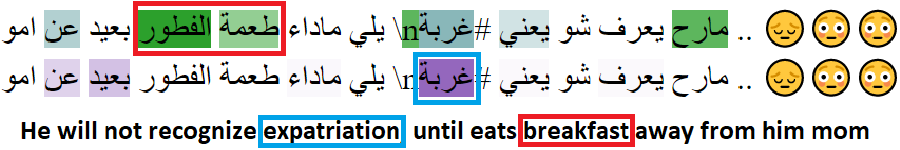}  
      \caption{Looking at different words}
      \label{fig:BetterEn3}
    \end{subfigure}    
    \caption{Examples of cross-lingual \enAraBERT~outperforming monolingual \arAraBERT}
    \label{fig:BetterEn}
\end{figure}

We analyzed 100 test examples with the greatest difference in Jaccard score between \arAraBERT~(trained on Arabic data) and \enAraBERT~(trained on translated English data), with only one of them producing the correct labels.  Specifically, we looked at 50 examples where \arAraBERT~was better than \enAraBERT~and 50 for the reverse.  We did the same for Spanish also with es-BetoBERT (trained on Spanish data) and en-BetoBERT (trained on translated English data).
Given the 100 examples that we analyzed, Figures \ref{fig:arabicErrorTypes} and \ref{fig:spanishErrorTypes} show the top error types along with their percentages for Arabic and Spanish respectively.  As the results show, the dominant error types depended greatly on which model performed better.  When \arAraBERT~performed better, the most common 3 \enAraBERT~error types (accounting for 80\% of the errors) were:
\begin{itemize}[leftmargin=*]
    \item The model missed an important word completely -- most likely due to translation.  In the example in Figure \ref{fig:BetterAr1}, \enAraBERT~missed the word corresponding to ``I hate'', which was written in dialectal Egyptian.
    \item Both models identified critical words, but weighted them differently.  As in Figure \ref{fig:BetterAr2},  \enAraBERT~and \arAraBERT~indicated that the word corresponding to ``disturbing'' is related to \textit{anger} (correct tag), but each gave it a different weight.
    \item The models had two different cultural interpretation to the same word. As in Figure \ref{fig:BetterAr3}, the word ``terrrorist'' was associated to \textit{disgust} by the \arAraBERT~while being associated with fear for by \enAraBERT.
\end{itemize}
Conversely, when \enAraBERT~performed better, the most common 3 \arAraBERT~error types (accounting for 76\% of the errors) were:
\begin{itemize}[leftmargin=*]
    \item Both models identified critical words, but weighted them differently.  As in Figure \ref{fig:BetterEn1}, \arAraBERT~and~ \enAraBERT~indicated that the word corresponding to ``best'' and ``laughter'' are related to \textit{love} (incorrect tag), with \arAraBERT~giving them much higher weight.
    \item The models produced tags that were different, but with matching sentiment (either positive or negative).  As in Figure \ref{fig:BetterEn2}, \arAraBERT~used ``sarcasm'' to assign \textit{sadness} and \textit{pessimism} tags, while \enAraBERT~used the same word to assign \textit{anger} and \textit{disgust} (correct tags).
    \item The models based their decisions on completely different words.  As in Figure \ref{fig:BetterEn3}, \arAraBERT~used ``breakfast'' to assign joy and \enAraBERT~used ``expatriation'' to assign \textit{sadness} (correct tag).
\end{itemize}

For Spanish, we omit details and figures due to space constraint and discuss only the important error types. When \esBetoBERT~outperformed~\enBetoBERT, the most prominent errors were somewhat different than those for Arabic. The top four error types, accounting for 70\% of the errors, were: the guesses of both models were different, but matched in sentiment; the models looked at different words; both models identified critical words, but weighted them differently; and some words were completely misclassified.  When \enBetoBERT~outperformed \esBetoBERT, the results were somewhat similar to those for Arabic, where the top 3 errors, accounting for 68\% of the cases, were: both models identified critical words, but weighted them differently; the models considered different words; and \esBetoBERT~ completely missed some critical words. \textcolor{black}{One important difference from Arabic 
is that errors stemming from cultural differences were less pronounced. 
}
In situations where a model completely missed a word or looked at the wrong word(s), such could potentially stem from translation errors.  For other cases where the models weighed words differently, such could be a result of differences in culture or language use. 


\section{Conclusion}
\label{discussion}
In this paper, we investigated the efficacy of cross-lingual models in the context of emotion detection compared to monolingual models. The effectiveness of such models are contingent on the similarity of how emotions are expressed across languages and cultures. We focused on using English emotion detection training data to train models that can tag Arabic/Spanish tweets using 11 different emotions with minimal drop in effectiveness compared to monolingual models. We presented three different cross-lingual approaches, namely using: i) multilingual models, ii) a translated training set, and ii) an automatically tagged parallel corpus. We experimented with these approaches individually and in combination with a varied set of classifiers. We also compared cross-lingual models with combined models that combine source and target language data.  
Using the translated English training set with fine-tuned contextual embeddings led to the best results for both Arabic and Spanish. Such cross-lingual models can avail the need to annotate language specific data, and show the transferability of emotions across languages and cultures.  
We also interpreted the results of the different models to understand why a cross-lingual model produced errors, while the monolingual model did not (or vice a versa). Our analysis shows that while data limitation is a challenge, translation errors and cultural difference can adversely affect cross-lingual models. 

\section{Bibliographical References}\label{reference}

\bibliographystyle{lrec2022-bib}
\bibliography{lrec2022-example,anthology}


\end{document}